\documentclass[letterpaper]{article} 
\usepackage{aaai24}  
\usepackage{times}  
\usepackage{helvet}  
\usepackage{courier}  
\usepackage[hyphens]{url}  
\usepackage{graphicx} 
\usepackage{natbib}  
\usepackage{caption} 
\usepackage{algorithm}
\usepackage{algorithmic}

\usepackage{subfigure}
\usepackage{multirow}
\usepackage{makecell}
\usepackage{amsmath,amssymb} 
\usepackage{booktabs}
\usepackage{multirow}
\usepackage{url}
\usepackage{subcaption}
\usepackage{array}
\usepackage{graphicx}
\usepackage{colortbl}  
\usepackage{xcolor}
\usepackage{soul} 
\usepackage{color, xcolor} 
\usepackage{tabularx}
\newcolumntype{L}[1]{>{\raggedright\let\newline\\\arraybackslash\hspace{0pt}}m{#1}}
\newcolumntype{C}[1]{>{\centering\let\newline\\\arraybackslash\hspace{0pt}}m{#1}}
\newcolumntype{R}[1]{>{\raggedleft\let\newline\\\arraybackslash\hspace{0pt}}m{#1}}
\usepackage{float}

\usepackage{amsmath,amsfonts,bm}

\def\eqref#1{equation~\ref{#1}}

\def\1{\bm{1}}

\DeclareMathAlphabet{\mathsfit}{\encodingdefault}{\sfdefault}{m}{sl}
\SetMathAlphabet{\mathsfit}{bold}{\encodingdefault}{\sfdefault}{bx}{n}

\newlength\savewidth
\newcommand{\tablestyle}[2]{\setlength{\tabcolsep}{#1}\renewcommand{\arraystretch}{#2}\centering\footnotesize}
\renewcommand{\paragraph}[1]{\vspace{1.25mm}\noindent\textbf{#1}}

\newcolumntype{x}[1]{>{\centering\arraybackslash}p{#1pt}}
\newcolumntype{y}[1]{>{\raggedright\arraybackslash}p{#1pt}}
\newcolumntype{z}[1]{>{\raggedleft\arraybackslash}p{#1pt}}

\newcommand{\app}{\raise.17ex\hbox{$\scriptstyle\sim$}}

\definecolor{deemph}{gray}{0.6}

\definecolor{baselinecolor}{gray}{.9}

\definecolor{dt}{HTML}{ADCAD8}
\definecolor{dt2}{HTML}{cddfe7}

\definecolor{defaultcolor}{HTML}{F7E0D5}

\let\cite\citep

\definecolor{citecolor}{HTML}{0071BC}
\definecolor{linkcolor}{HTML}{ED1C24}
\usepackage{newfloat}
\usepackage{natbib}
\usepackage{listings}
\DeclareCaptionStyle{ruled}{labelfont=normalfont,labelsep=colon,strut=off} 
\floatstyle{ruled}
\newfloat{listing}{tb}{lst}{}
\floatname{listing}{Listing}
\title{Robustness-Guided Image Synthesis for Data-Free Quantization}
\author{
    Jianhong Bai$^{1}$, 
    Yuchen Yang$^{1}$,
    Huanpeng Chu$^{2}$,
    Hualiang Wang$^{3}$,
    Zuozhu Liu$^{1}$,\\
    Ruizhe Chen$^{1}$,
    Xiaoxuan He$^{1}$,
    Lianrui Mu$^{1}$,
    Chengfei Cai$^{4}$,
    Haoji Hu$^{1}$\thanks{Corresponding author.}
}
\affiliations{
    \textsuperscript{\rm 1}Zhejiang University\\
    \textsuperscript{\rm 2}Kuaishou Technology
    \\ \textsuperscript{\rm 3}The Hong Kong University of Science and Technology
    \\ \textsuperscript{\rm 4}Tencent Data Platform\\
    jianhongbai@zju.edu.cn
}

\usepackage{bibentry}

\begin{document}

\maketitle

\begin{abstract}
Quantization has emerged as a promising direction for model compression. Recently, data-free quantization has been widely studied as a promising method to avoid privacy concerns, which synthesizes images as an alternative to real training data. Existing methods use classification loss to ensure the reliability of the synthesized images. Unfortunately, even if these images are well-classified by the pre-trained model, they still suffer from low semantics and homogenization issues. Intuitively, these low-semantic images are sensitive to perturbations, and the pre-trained model tends to have inconsistent output when the generator synthesizes an image with poor semantics. To this end, we propose Robustness-Guided Image Synthesis (RIS), a simple but effective method to enrich the semantics of synthetic images and improve image diversity, further boosting the performance of downstream data-free compression tasks. Concretely, we first introduce perturbations on input and model weight, then define the inconsistency metrics at feature and prediction levels before and after perturbations. On the basis of inconsistency on two levels, we design a robustness optimization objective to enhance the semantics of synthetic images. Moreover, we also make our approach diversity-aware by forcing the generator to synthesize images with small correlations in the label space. With RIS, we achieve state-of-the-art performance for various settings on data-free quantization and can be extended to other data-free compression tasks.
\end{abstract}

\section{Introduction}
\label{sec:intro}




\begin{figure}[t]
	\centering
	\includegraphics[width=0.40\textwidth]{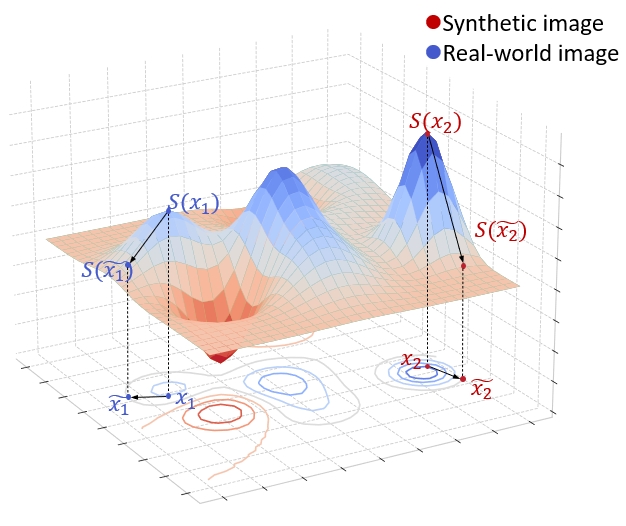}
	\caption{Visualization of the loss landscape. Note that we reverse the y-axis for visual convenience.}
	\label{fig:loss_landscape}
\end{figure}
Recently, deep neural networks have achieved great accomplishment in many areas, including computer vision \cite{he2016deep,girshick2015fast,sandler2018mobilenetv2} and natural language processing \cite{zia2019long,mikolov2010recurrent,devlin2018bert}. Despite their satisfactory performance, the huge number of parameters and high computational cost prevent them from being deployed to edge computing devices. Quantization, which converts parameters from full-precision to low-precision, has become a promising method for model compression. A large number of methods~\cite{liu2020reactnet,liu2018bi,fan2020training,zhuang2018towards} minimize the quantization error through training data and achieve great performance in low-bit quantization. However, accessibility to the original training data during the quantization process is not always possible due to privacy and security reasons, especially in medical and industrial scenarios. Data-driven methods would fail in this case.

Therefore, data-free quantization has been proposed for compression under privacy protection. Among some excellent works \cite{nagel2019data,banner2018aciq,cai2020zeroq, nagel2019data}, generative methods \cite{choi2021qimera,xu2020generative,2021AutoReCon, AIT,qian2023adasg,qian2023adaptive,2023hast, shang2023causal, choi2020data} have drawn much attention due to their great performance. These methods synthesize fake data through generators and use them to calibrate or fine-tune the quantified model. Hence, the quality of synthesized images influences the quantization performance greatly.

Beyond various image priors (e.g., match BN statistics~\cite{cai2020zeroq}) being used during image synthesis, most of the existing methods feed the synthesized images to the pre-trained model, then minimize the cross-entropy loss to guarantee the semantics of images \cite{xu2020generative,zhang2021diversifying}. Nevertheless, even if the images are well-classified, they still suffer from low semantics, limiting the performance of downstream model compression tasks. 
The phenomenon can be explained in two aspects. 
On one hand, studies \cite{goodfellow2014explaining,moosavi2016deepfool} in model attack show that deep neural networks can be easily misled by adding adversarial perturbations. Therefore, it is unsurprising that the pre-trained model has high confidence in these noise-like data. On the other, the lack of a discriminator in data-free scenarios greatly limits the performance of the generator, and it only needs to match the image and class priors, but ignores the semantics of synthesized images. As a result, classification loss is insufficient to guide the generator in synthesizing samples with high semantics.

To this end, we manage to enrich the semantics of synthetic images in our work. Our intuition is that the low-quality images synthesized by existing methods are easily hampered, while real-world data are more robust towards perturbations due to their rich semantic information. In other words, the pre-trained model tends to have consistent representation and prediction which may not be easily disturbed for semantically abundant images (e.g., real-world data). To verify our hypothesis, we conduct a motivational experiment, as shown in Fig. \ref{fig:loss_landscape}. Concretely, we visualize the loss landscape of GDFQ \cite{xu2020generative}, one of the SOTA methods in data-free quantization, and we plot the training data (real-world image) and synthesized image on the loss landscape as $\boldsymbol{x}_1$ and $\boldsymbol{x}_2$ respectively. Then, we apply input and weight perturbations to both images and denote the disturbed images as $\Tilde{\boldsymbol{x}}_1$ and $\Tilde{\boldsymbol{x}}_2$. It's observed that the real-world image has a relatively smooth neighborhood in the loss landscape, while the loss of synthetic image changes dramatically nearby, implying the representation or prediction could have a large discrepancy before and after perturbations. 

Based on the above observation, we propose a simple but effective method called Robustness-Guided Image Synthesis (RIS) to enrich the semantics of synthesized images, and further boost downstream data-free model compression tasks. 

Concretely, we first conduct motivational experiments to verify that these low semantic images synthesized by existing methods are sensitive toward perturbations, which is different from real data with rich semantic information. Then, we explicitly model the inconsistency at feature and prediction levels before and after different kinds of perturbations as image robustness, and further design a robustness optimization objective for training the generator. The proposed robustness loss significantly improves the semantics of the synthetic images by forcing a smooth neighborhood in the loss landscape, as visualized in Fig. \ref{fig:loss_landscape}. On the other, we also alleviate the image homogenization problem \cite{zhang2021diversifying} via formulating an optimization problem and replacing the ont-hot label set with multiple soft labels with minimal correlation. With RIS, the Fréchet Inception Distance (FID)~\cite{FID} and Inception Score (IS)~\cite{IS} of synthetic images outperform the baseline with an improvement of \textbf{80.13} and \textbf{22.44} on ImageNet~\cite{krizhevsky2012imagenet}.

Experiments on a variety of pre-trained models and datasets show RIS consistently achieves significant performance improvement. Moreover, our method is not limited to quantization, which can be extended to other data-free scenarios such as data-free knowledge distillation \cite{hinton2015distilling}.

We make the following major contributions. \textbf{1)} We identify that the images synthesized by existing methods are more sensitive toward perturbations than natural images, leading to the low-semantic and limiting the performance of downstream tasks. \textbf{2)} We propose the Robustness-Guided Image Synthesis (RIS) scheme, a simple but effective method to enrich the semantics and improve the diversity of synthetic images. \textbf{3)} We conduct extensive experiments, showing that the proposed RIS outperforms various existing data-free quantization methods by a large margin, and can be further extended to data-free knowledge distillation.

\section{Related Work}
\subsection{Data-Free Model Compression}
\textbf{Knowledge Distillation} Early works in data-free scenarios focus on knowledge distillation \cite{hinton2015distilling}, which devise different regularizations for learning image priors \cite{lopes2017data,nayak2019zero,zhu2021data,yu2023data,binici2022preventing,hao2022cdfkd}. They can be roughly divided into three categories: 
\textbf{1) synthesis from noise.}
Lopes et al. \cite{lopes2017data} utilize the activation records (i.e., means and covariance), which are restored as metadata for reconstructing training data.
Nayak et al. \cite{nayak2019zero} model the softmax space of the teacher as a Dirichlet distribution and craft data from the parameters of the teacher.
DeepInversion \cite{yin2020dreaming} combines the image prior presented by \cite{mordvintsev2015inceptionism} and aligns the BN statistics of the real ones. Although these methods can obtain high semantic images, it has the drawback of huge computational costs because each batch of synthetic images has to optimize from the beginning.
\textbf{2) reconstruction with a generator.}
Represented by DAFL \cite{chen2019data}, these methods synthesize images from a generator. DAFL exploits cross-entropy loss as the class prior and maximum activation as the semantic prior. Luo et al. \cite{luo2020large} use multi-generators and further extend to the large-scale dataset i.e. ImageNet \cite{krizhevsky2012imagenet}.
\textbf{3) adversarial exploration.}
ZSKT \cite{micaelli2019zero} and DFAD \cite{fang2019data} train an adversarial generator to search for images where the prediction of the student poorly matches the teacher's prediction. However, these methods are sensitive to hyperparameters.

\textbf{Quantization} Most data-free quantization methods share a similar pipeline with data-free knowledge distillation: first synthesize images, then utilize them as surrogates for the original training data. ZeroQ \cite{cai2020zeroq}, as the pioneer of the generative method for quantization, synthesizes data that match the statistics of BN \cite{ioffe2015batch} layers. GDFQ \cite{xu2020generative} further uses the cross-entropy loss to ensure the synthesized images can be classified by the pre-trained network correctly. Based on the GDFQ framework, DSG \cite{zhang2021diversifying} optimizes the diversity of synthesized images by relaxing BN statistics constraint, while Qimera \cite{choi2021qimera} focuses on generating samples nearing the classification boundary by using superposed latent embeddings. 
AutoReCon \cite{2021AutoReCon} first searches for an optimized neural architecture to reconstruct the generator. 
AIT \cite{AIT} argues that the classification loss and KL divergence have gradient confliction thus excluding the cross-entropy loss and proposes gradient inundation to solve infrequent updates of the quantized model.
IntraQ \cite{zhong2022intraq} imitates real data by generating heterogeneous synthetic images.
HAST \cite{2023hast} solves real data degradation by synthesizing hard samples and further promoting sample difficulty while training models.
AdaSG \cite{qian2023adasg} enhances image adaptability while ensuring model accuracy by rethinking data-free quantization as a zero-sum game between the generator and the quantized network.
AdaDFQ \cite{qian2023adaptive} further improves it by optimizing the margin between the lower and upper boundaries defined by disagreement and agreement samples.
These approaches consider the distribution of synthesized data at the statistical level but ignore the critical issue of generation quality. All these methods suffer from synthesizing unrealistic images with low semantics.

\subsection{Image Robustness}

Robustness~\cite{robust1} is defined as the property of a procedure that renders the answers it gives insensitivity to departures, which is used to describe systems, models, or images. Image robustness~\cite{robust2,robust3} is a concept in signal processing that means the image still has a certain degree of fidelity after various signal processing or attacks. In the field of watermarking, a group of methods design robust watermarks to resist the transformations of images \cite{watermark}. Image steganography communicates secret data by adding undetectability and robustness signal to the original image \cite{steganography}.
In model defense \cite{papernot2016towards}, a few methods exploit the prediction inconsistency to detect adversarial examples \cite{featuresqueezing,meng2017magnet,defence1}, which share the same spirits with our work. 
Xu et al. \cite{featuresqueezing} introduce pre-processing the input images, then calculating the prediction shift score to separate adversarial samples from the natural ones. In this paper, we also transform the images and make use of the inconsistency. The biggest difference is that we explicitly model the robustness at feature and prediction levels, and use it as a guidance for image synthesis.

\section{Method}

\subsection{Generative Data-Free Quantization}
Generative data-free quantization methods, pioneered by \cite{xu2020generative}, have drawn great attention due to their excellent performance and efficiency, which employ a generator $G$ to fit real training data distribution. With synthesized data, the quantized model is fine-tuned by mimicking the behavior of the pre-trained model $\mathcal{M}$ through knowledge distillation.
Take GDFQ \cite{xu2020generative} as an example, the generator synthesize fake data $\boldsymbol{x}$ from the Gaussian noise $\textbf{z}$, conditional on the one-hot label $y$:
\begin{equation}\label{eq:generate}
	\begin{aligned}
		\boldsymbol{x} = G(\textbf{z}\mid y),       \qquad \textbf{z} \sim  \mathcal{N}(0,1)
	\end{aligned}
\end{equation} 

The generated image is then classified by the pre-trained model $\mathcal{M}$, which is composed of a feature extractor $f$ and a classifier $g$, and update the generator by the classification loss:

\begin{equation}\label{CE_loss}
	\begin{aligned}
		\mathcal{L}_{\text{CE}}(G) = \mathbb{E}_{\textbf{z},y} \left[ \mathrm{CE}(g(f(\boldsymbol{x})),y)\right],
	\end{aligned}
\end{equation} 

Meanwhile, mean square error (MSE) loss is used to align the mean and variance at BN layers:
\begin{equation}\label{BNS_loss}
	\mathcal{L}_{\text{BNS}}(G) = \sum_{l=1}^{L}\Vert{\boldsymbol{\mu}_l^r-\boldsymbol{\mu}_l}\Vert_2^2 + \Vert{\boldsymbol{\sigma}_l^r-\boldsymbol{\sigma}_l}\Vert_2^2,
\end{equation} 
where $\mu_l^r$ and $\sigma_l^r$ refer to the mean and variance of the synthesized images at the $l$-th BN layer, $\mu_l$ and $\sigma_l$ is the mean and variance stored in the pre-trained model. The overall objective for generator $G$ is:
\begin{equation}\label{GDFQ_loss}
	\mathcal{L}_{\text{GDFQ}}(G) = \mathcal{L}_{\text{CE}}(G) + \alpha \mathcal{L}_{\text{BNS}}(G).
\end{equation}

Based on the GDFQ framework, a few methods attempt to improve the quality of generated data from different aspects. DSG~\cite{zhang2021diversifying}, which aims to improve the diversity of synthesized images, modifies $\mathcal L_{BNS}(G)$ term by adding slack variables and designing layerwise sample enhancement. While Qimera~\cite{choi2021qimera} considers images near the decision boundary to be more helpful for data-free quantization.
These approaches improve the samples at the distribution level, but ignore the critical point \textemdash the semantic and quality of synthesized images. 

\subsection{Semantic Enhancement with Input and Weight Perturbations}
\label{sec_semantic}
In this section, we delve into enhancing the semantic information of the synthetic images and further boosting the performance of data-free model compression tasks (e.g., quantization). Concretely, we first propose to involve perturbations on both input and weight levels, then explicitly define a robustness metric via the model inconsistency on feature representation and predicted distribution after perturbations. Finally, we design an additional optimization objective for supervising the generator based on the robustness metric.

\subsubsection{Perturbations on Input and Model Weight} 
\label{sec_perturb}

\begin{figure*}[t]
	\centering
	\includegraphics[width=0.90\linewidth]{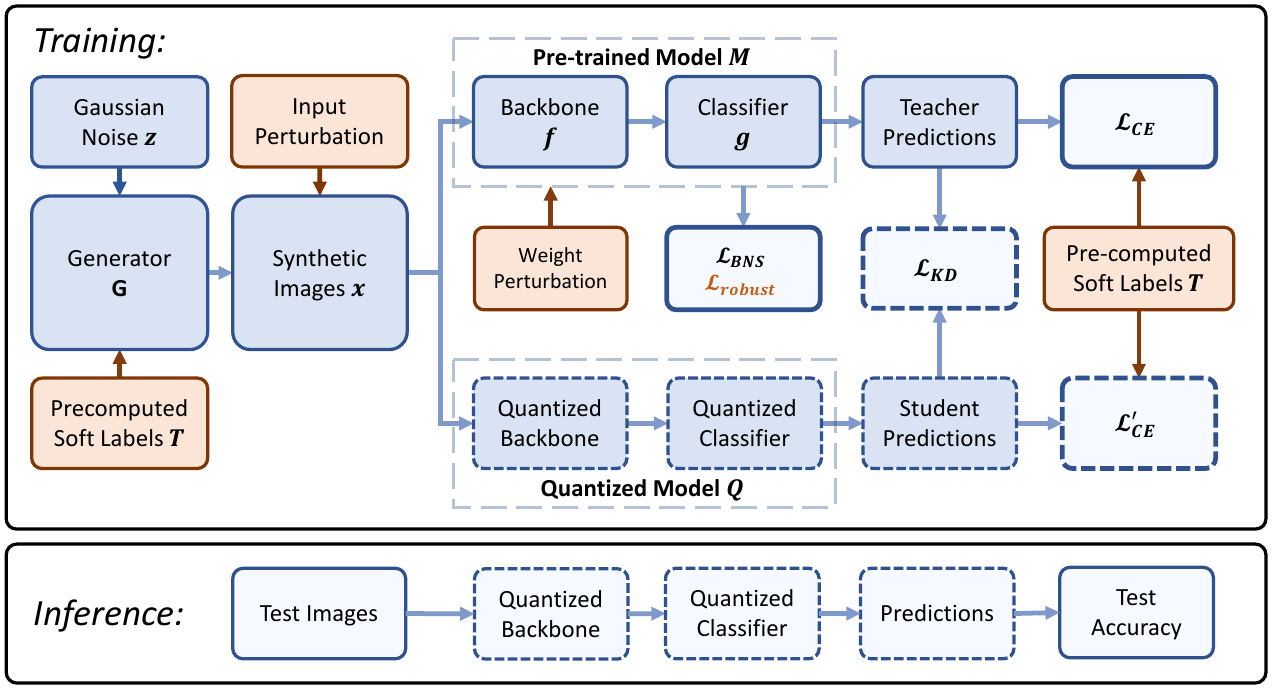}
	\caption{The pipeline of RIS. The solid line boxes refer to the teacher (pre-trained) model and losses for updating the generator $G$, while the dashed line boxes are the quantized model and its loss function. Proposed components are denoted as orange.}
	\label{fig:pipeline}
\end{figure*}

Recall in Fig. \ref{fig:loss_landscape}, we verify that low semantic images are more sensitive towards perturbations. Hence, our core idea is to explicitly optimize the generator to synthesize images that the teacher model has consistent outputs before or after the perturbations. To this end, the first step of our method is to introduce perturbations from input and model weight. 
For input perturbation, we implement it with several data augmentation strategies. As shown in Figure \ref{fig:pipeline}, the original images $\boldsymbol{x}$ as well as the augmented ones $\mathcal{A}(\boldsymbol{x})$ are fed into the full-precision (teacher) model and we optimize the generator with a consistency loss which will introduce in the following sections. It's worth noting that input perturbations can be instantiated as \textbf{any} differentiable data augmentation method (see Appendix for details). \footnote{Since we need to back-propagate the gradient to update the generator, the augmentation strategy must be differentiable.}

On the other, we further introduce perturbation on weight as complementary. Given the teacher model $({g} \circ f)$ with parameters $\boldsymbol{w}$, we add a perturbation term $\boldsymbol{v}$ to the original weight, i.e., $\boldsymbol{w}^{\prime} = \boldsymbol{w} +  \boldsymbol{v}$. As shown in Figure \ref{fig:pipeline}, we explore three different weight perturbation strategies: a), we instantiate the perturbation term as the Gaussian noise with mean $\mu$ and deviation $\sigma$: $\boldsymbol{v} \sim \mathcal{N}(\mu, \sigma ^ 2)$; b), we conduct experiments with adversarial weight perturbation (similar with several studies in model defense \cite{adversarial1, adversarial2, AWP}), i.e., $ \boldsymbol{v} = \gamma \, \frac{\nabla_w \mathrm{CE}[(g \circ f)_w (\boldsymbol{x}_i), y]}{\nabla_w \Vert \mathrm{CE}[(g \circ f)_w (\boldsymbol{x}_i), y] \Vert} \, \Vert \boldsymbol{w} \Vert$, which is also shown as effective in experiments; c), Dropout \cite{dropout} as a widely-used technique to enhance model generalization and reducing overfitting, can also regard as a weight perturbation strategy where $\boldsymbol{v}$ is $-\boldsymbol{w}$ for discarded neurons and zero for others.

\subsubsection{Robustness Modeling via Inconsistency} 
\label{sec_robust_modeling}

In this section, we explicitly model image robustness by the maximum value of inconsistency at the level of feature and prediction after $n$ perturbations. At the feature level, we exploit the cosine distance to measure the variation of features:

\begin{equation}
	{\mathcal{R}_{f}(\boldsymbol{x}) = \max \limits_{1\le i\le n}  \frac{\left<f(\boldsymbol{x}), \mathcal{A}_i(f(\boldsymbol{x}))\right> }{\Vert f(\boldsymbol{x}) \Vert \cdot \Vert \mathcal{A}_i (f(\boldsymbol{x})) \Vert}} ,
	\label{eq:df}
\end{equation}
where $\boldsymbol{x}$ refers to the synthesized image, $\mathcal{A}_i$ is the $i$-th different perturbation strategy, $f(\cdot)$ denotes the operation of extracting the features in the pre-training model, and $\mathcal{R}_{f}(\boldsymbol{x})$ denotes the inconsistency of the synthetic image $\boldsymbol{x}$ on feature level. Since the details or textures of the synthetic image may not be held after perturbations, we utilize the feature embedding of the last layer which indicates the high-level semantics (should be consistent towards perturbations). 

At the prediction level, we use the $L_1$ distance to quantify the variation:

\begin{equation}
	\mathcal{R}_{p}(\boldsymbol{x}) = \max \limits_{1\le i\le n} \Vert g( f(\boldsymbol{x})) -  \mathcal{A}_i(g(f(\boldsymbol{x}))) \Vert_1,
	\label{eq:dp}
\end{equation}
where $g(\cdot)$ refers to extracting the prediction vector from the pre-trained model, which is commonly a single fully-connected layer. Note that We choose the $max$ operator based on the observation that real images are consistently stable under a variety of perturbations (detailed in Appendix B), while the synthesized ones fluctuate greatly. Hence, minimizing the max value of inconsistency can provide strong regularization to the generator.

\subsubsection{Robustness-guided Image Synthesis}

Finally, we formulate the optimization objective based on the two levels of inconsistency $\mathcal{R}_{f}$ and $\mathcal{R}_{p}$ introduced above:

\begin{equation}
\small
	\mathcal{L}_{\text{robust}}(G) = \mathbb{E}_{\boldsymbol{x}} [\underbrace{\max(\mathcal{R}_{f}(\boldsymbol{x})-\theta_f,0)+\beta\max(\mathcal{R}_{p}(\boldsymbol{x})-\theta_p,0)}_{\mathcal{R}(\boldsymbol{x})}],
	\label{eq_robust}
\end{equation}
where $\theta_f$ and $\theta_p$ denote the threshold for the distance of model output before and after perturbations at feature and prediction levels, respectively. $\beta$ is the weighting coefficient to balance the consideration of feature and prediction. $\mathcal{R}(\boldsymbol{x})$ represents the robustness of image $\boldsymbol{x}$. We regard the image as robust towards augmentations only when the variation of feature and prediction is lower than the threshold simultaneously. Note that a lower value of $\mathcal{R}(\boldsymbol{x})$ indicates high robustness towards data augmentation. Thus, it can be directly integrated into the existing loss of the generator as $\mathcal{L}_{\text{robust}}$.

A significant challenge is determining the values of $\theta_f$ and $\theta_p$ without any real data access. Fortunately, the noise data has a similar performance to the synthesized images (detailed in Appendix B), it can be used as a representation of low-semantic images. Therefore, we initialize 1000 samples $\boldsymbol{x}_{\text{noise}}$ from $\mathcal{N}(0,1)$, which are further fed into the teacher model to compute $\mathcal{R}_{f}(\boldsymbol{x}_{\text{noise}})$ and $\mathcal{R}_{p}(\boldsymbol{x}_{\text{noise}})$. We set the thresholds to the $\epsilon$ percentile of these values:
\begin{equation}
	\theta_f = \mid \mathcal{R}_{f}(\boldsymbol{x}_{\text{noise}})\mid_\epsilon, \qquad \theta_p = \mid \mathcal{R}_{p}(\boldsymbol{x}_{\text{noise}})\mid_\epsilon.
	\label{equ:theta}
\end{equation}

When the $\epsilon$ becomes larger, the loss function in Eq.~\ref{eq_robust} has less tolerance for non-robustness samples. The default value of $\epsilon$ is set as 0.1. 
As a result, The low semantic images are eliminated by optimizing their robustness towards both input and weight perturbations, enabling the generator to synthesize images with high semantics (detailed in Appendix C).

\subsection{Diversity-aware Image Synthesis}
In the above section, we force the generator to synthesize images with rich semantic information via the robustness constraint in Eq. \ref{eq_robust}. While another important indicator for evaluating generative models is the diversity of synthesized images. A recent work \cite{what_make} suggests that training images with small correlation results in better performance on knowledge distillation scenarios. Hence, we propose explicitly improving diversity by replacing $C$ one-hot labels in ACGAN with $N$ ($N > C$) soft labels with the smallest correlation.

Concretely, the original labels in GDFQ \cite{xu2020generative} can be viewed as an identity matrix $I \in \mathbb{R}^{C \times C}$, and we replace it with soft labels $T \in \mathbb{R}^{N \times C}$. The intuition behind this is that there are always multiple prototypes within one semantic class, e.g., the class `cat' could contain different breeds of cats or cats with different backgrounds. On the other, we should avoid any two label vectors being too similar since it leads to the issue of homogenization for synthetic images. To this end, we formulate it as an optimization problem:
\hspace{-0.5cm}
\begin{equation}
    \begin{aligned}
        &{\textbf{Minimize:}}\\
        & \qquad \sum_{i=1}^N \sum_{j=1}^N 1 \, / \, \text{dis}(T_i, T_j). \\
        &{\textbf{Subject to:}}\\
            & \qquad \begin{cases}
            \sum_{j=1}^N T_{ij} = 1 \quad  &\forall j = 1, \cdots, N \\
            T_{ij} \geq 0 \quad & \forall i,j = 1, \cdots, N 
            \end{cases}
    \end{aligned}
    \label{equ:opt}
\end{equation}
where $T$ is the modified target matrix for training the generator, $\text{dis}(\cdot)$ refers to a distance metric in Euclidean space. Since each row of $T$ represents a probability distribution, the sum of each row should be 1, and each term greater than 0. We solve the defined optimization problem in Eq. \ref{equ:opt} with SGD before training, and supervised the generator with the calculated soft labels in $T$. In this way, the generator is forced to synthesize diverse samples with multiple prototypes within a semantic class, and those prototypes are explicitly having minimum correlations with each other, which is empirically effective in experiments.
The whole process of the RIS scheme is summarized in Algorithm \ref{alg}.

\begin{algorithm}[t]
	\caption{The synthesis process of our RIS
		scheme.}
	\textbf{Input}: Pre-trained model $\mathcal{M}$, training epochs $T_e$, label matrix $T$, hyper-parameters $\alpha$, $\beta$.
	\begin{algorithmic}[1] 
		\STATE Initialize $\boldsymbol{x}_\text{{noise}}$ from Gaussian distribution $\mathcal{N}$(0, 1).
		\STATE Feed $\boldsymbol{x}_\text{{noise}}$ into $\mathcal{M}$ to compute $\mathcal{R}_f(\boldsymbol{x}_\text{{noise}})$ and $\mathcal{R}_p(\boldsymbol{x}_\text{{noise}})$  based on Eq.~\ref{eq:df} and Eq.~\ref{eq:dp}.
		\STATE Compute $\theta_{f}$ and $\theta_{p}$ according to $\epsilon$.
		\FOR{$epoch = 1:T_e$}
		\STATE Sample random noise \textbf{z} $\sim$ $\mathcal{N}$(0, 1) and soft label $\tilde{\boldsymbol{y}}$ in $T$.
		\STATE Generate fake image $\boldsymbol{x}$ using Eq.~\ref{eq:generate}.
		\STATE Input perturbations: Obtain the perturbed images \{$\mathcal{A}_{i}(\boldsymbol{x})\}_{i=1}^{n}$ through $n$ data augmentations.
  	\STATE Weight perturbations: Obtain the perturbed teacher model $\{ \mathcal{M}^{\prime}_j\}_{j=1}^{m}$ through $m$ different weight perturbation strategies introduced in Section \ref{sec_perturb}.
		\STATE Feed the original image and the perturbed ones $\{\boldsymbol{x} \cup \{\mathcal{A}_{i}(\boldsymbol{x})\}_{i=1}^{n}\}$ into $\{ \mathcal{M} \cup \{ \mathcal{M}^{\prime}_j\}_{j=1}^{m}\}$. 
        \STATE Calculate $D_f(I)$ and $D_p(I)$ with Eq.~\ref{eq:df} and Eq.~\ref{eq:dp}.
		\STATE Calculate the cross-entropy loss $\mathcal{L_{\text{CE}}}$ with the corresponding sampled label $\tilde{\boldsymbol{y}}$ and BN statistic loss $\mathcal{L_{\text{BNS}}}$ via Eq.~\ref{CE_loss} and Eq.~\ref{BNS_loss} respectively.
		\STATE Calculate the proposed $\mathcal{L}_{\text{robust}}$ through Eq.~\ref{eq_robust}.
		\STATE Update the generator $G$ by minimizing $\mathcal{L_{\text{RIS}}} = \mathcal{L_{\text{CE}}} + \alpha \mathcal{L_{\text{BNS}}} + \beta \mathcal{L}_{\text{robust}}$.
		\ENDFOR
        
	\end{algorithmic}
	\label{alg}

\end{algorithm}	

\section{Experimental Results}
\subsection{Experiment Implementation}
Following previous work, we evaluated the proposed RIS on CIFAR-10/100 \cite{krizhevsky2009learning} and ImageNet  \cite{krizhevsky2012imagenet}. 
In order to facilitate comparison with existing methods, we choose ResNet-20 \cite{he2016deep} for CIFAR-10/100, and ResNet-18, ResNet-50, and MobileNetV2 \cite{sandler2018mobilenetv2} for ImageNet. 
In addition, due to the low semantic images being little helpful to the quantized model, we set several epochs to warm up the generator, i.e., only updating the generator $G$ at the beginning and the quantized model $Q$ is not updated. More training details can be found in Appendix D.

\subsection{Quantization Results on Various Baselines}
To verify the effectiveness and versatility of RIS, we employ our method on various widely-used network architectures based on the four advanced generative methods: GDFQ \cite{xu2020generative}, DSG \cite{zhang2021diversifying}, AutoRecon \cite{2021AutoReCon}, AIT \cite{AIT}.
The results are displayed in Table~\ref{tab:cifa_imagenet}. Note that W$n$A$m$ means $n$-bit quantization for weights and $m$-bit quantization for activations. We report top-1 accuracy for each experiment. Note that AIT is a plug-in approach, AIT in the table refers to ARC-based AIT. The observations can be summarized as \textbf{1)} RIS outperforms the baselines in almost all settings, especially on low bit-width. The only slight degradation occurs in DSG with the CIFAR-10 dataset which is already close to the teacher accuracy thus leaving little room for improvement. \textbf{2)} RIS leads to extraordinary improvements in the quantized ResNet-50 model, increasing the accuracy of the 4-bit quantized GDFQ model by 10.04\% and the DSG model by 8.72\%. In particular, while the AIT + ARC original accuracy rate had reached 68.27\%, RIS still brought an increase of 3.27\%, making the prediction accuracy reach an astonishing 71.54\%.

\begin{table*}[h]
	\begin{center}


        \setlength\tabcolsep{8.5pt}
  \begin{tabular}{lll|ll|llll}
			\toprule
			\multirow{5}{*}{Method} & \multicolumn{2}{c}{CIFAR-10} & \multicolumn{2}{c}{CIFAR-100} & \multicolumn{4}{c}{ImageNet}  \\ 
   \cmidrule(r){2-3} \cmidrule(r){4-5} \cmidrule(r){6-9}
                & \multicolumn{2}{c}{ResNet-20} & \multicolumn{2}{c}{ResNet-20}                     & \multicolumn{2}{c}{ResNet-18} & \multicolumn{2}{c}{ResNet-50}\\
			&\multicolumn{2}{c}{(93.89)}&\multicolumn{2}{c}{(70.33)}&\multicolumn{2}{c}{(71.47)} & \multicolumn{2}{c}{(77.73)}\\ \cmidrule(r){2-3} \cmidrule(r){4-5} \cmidrule(r){6-7} \cmidrule(r){8-9}
			
			& W4A4 & W5A5 & W4A4 & W5A5 & W4A4 & W5A5 & W4A4 & W5A5\\ \midrule
			
			GDFQ &90.25 &93.38   &63.39 &66.12&60.60&68.40&52.12&71.89\\ 
			+RIS&$\textbf{91.04}^{{\textbf{+0.79}}}$  
                &$\textbf{93.59}^{{\textbf{+0.21}}}$ 
                &$\textbf{65.50}^{{\textbf{+2.11}}}$ 
                &$\textbf{69.06}^{{\textbf{+2.96}}}$ 
                &$\textbf{62.81}^{{\textbf{+2.21}}}$ 
                &$\textbf{69.77}^{{\textbf{+1.37}}}$ 
                &$\textbf{62.16}^{{\textbf{+10.04}}}$ 
                &$\textbf{75.24}^{{\textbf{+3.35}}}$ \\ \midrule
			DSG &91.05 & \textbf{93.56}  &63.42 &67.25&61.58 & 69.53&54.68&72.25 \\
			
			+RIS &$\textbf{92.59}^{{\textbf{+1.54}}}$      
                & 93.50$^{{\textbf{-0.06}}}$  
                &$\textbf{65.99}^{{\textbf{+2.57}}}$ 
                &$\textbf{69.55}^{{\textbf{+2.30}}}$ 
                &$\textbf{64.59}^{{\textbf{+3.01}}}$ 
                &$\textbf{69.84}^{{\textbf{+0.31}}}$ 
                &$\textbf{63.40}^{{\textbf{+8.72}}}$ 
                &$\textbf{75.18}^{{\textbf{+2.93}}}$\\ \midrule
			ARC &88.55 &92.88 &62.76 &68.40 &61.32&68.88&64.37&74.13\\ 
			+RIS  
                &$\textbf{91.44}^{{\textbf{+2.89}}}$ 
                &$\textbf{93.49}^{{\textbf{+0.61}}}$ 
                &$\textbf{63.82}^{{\textbf{+1.06}}}$ 
                &$\textbf{69.15}^{{\textbf{+0.75}}}$ 
                &$\textbf{63.58}^{{\textbf{+2.26}}}$ 
                &$\textbf{69.26}^{{\textbf{+0.38}}}$ 
                &$\textbf{68.40}^{{\textbf{+4.03}}}$ 
                &$\textbf{75.40}^{{\textbf{+1.27}}}$ \\ \midrule
			AIT  &87.93${}^{\ast}$ &92.89  &61.05 &68.40&65.73&70.28&68.27&76.00\\ 
			+RIS &$\textbf{89.84}^{{\textbf{+1.91}}}$ 
                &$\textbf{93.23}^{{\textbf{+0.34}}}$ 
                &$\textbf{63.51}^{{\textbf{+2.46}}}$ 
                &$\textbf{68.94}^{{\textbf{+0.54}}}$ 
                &$\textbf{67.55}^{{\textbf{+1.82}}}$ 
                &$\textbf{70.59}^{{\textbf{+0.31}}}$ 
                &$\textbf{71.54}^{{\textbf{+3.27}}}$ 
                &$\textbf{76.36}^{{\textbf{+0.36}}}$\\ 
			\bottomrule
		\end{tabular}
	\end{center}
 		\caption{Results on CIFAR-10/100 and ImageNet with various baseline methods. ``${\ast}$" denotes our re-implementation.}
   		\label{tab:cifa_imagenet}
\end{table*}
\par
\noindent

\subsection{Ablation Studies}
\label{Sec: ablation}

 \begin{table*}[h]


	\begin{center}
            \subfigure[{Input Perturbation.}
		              \label{tab:ablation_input}]{
						  \tablestyle{4pt}{1.05}

    						\begin{tabular}{L{2cm}C{1.3cm}C{1.3cm}}
							\toprule
                                \makecell[l]{Augmentation\\Strategy}&\makecell[c]{Top-1\\Accuracy}&\makecell[c]{Top-5\\Accuracy} \\
							\midrule
                                Baseline & 63.39&87.59\\
							Noise  &  64.38&87.87\\
							Translation &  64.42&87.91\\
							Resize & 64.40& 88.08\\
                                 Random Select  & \textbf{64.88}& \textbf{88.21}\\
							\bottomrule
						\end{tabular}}
      \hspace{0.017\linewidth}
            \subfigure[{Weight Perturbation.}
                        \label{tab:ablation_weight}]{
						  \tablestyle{4pt}{1.05}

    						\begin{tabular}{L{2cm}C{1.3cm}C{1.3cm}}
							\toprule
                                \makecell[l]{Weight\\Perturbation}&\makecell[c]{Top-1\\Accuracy}&\makecell[c]{Top-5\\Accuracy}\\
							\midrule
                                Baseline & 63.39&87.59\\
							 Gaussian Noise & \textbf{64.48}  & \textbf{87.63} \\
							Adversarial & 63.65 & 86.79\\
							Dropout & 64.14  &87.57\\

                                 & &\\
							\bottomrule
						\end{tabular}}
      \hspace{0.017\linewidth}
                  \subfigure[{Components of RIS}
                        \label{tab:ablation_component}]{
						\tablestyle{4pt}{1.05}
    						\begin{tabular}{L{2cm}C{1.3cm}C{1.3cm}}
							\toprule
                                Strategy &\makecell[c]{Top-1\\Accuracy}&\makecell[c]{Top-5\\Accuracy}\\
							\midrule
                                Baseline & 63.39&87.59\\
							Input Perturb& 64.38 &88.21\\
							Weight Perturb& 64.48 & 87.63\\
							Soft Label & 63.99 & 87.61\\
							 All & \textbf{65.50}& \textbf{88.44}\\
							\bottomrule
						\end{tabular}}\\
            \subfigure[{Hyper-parameter $\varepsilon$}
                        \label{tab:ablation_epsilon}]{
						  \tablestyle{4pt}{1.05}
    						\begin{tabular}{L{2cm}C{1.3cm}C{1.3cm}}
							\toprule
                                $\varepsilon$ &\makecell[c]{Top-1\\Accuracy}&\makecell[c]{Top-5\\Accuracy} \\
							\midrule
                                0 & 64.95&87.79\\
                                0.05 &  65.05&88.21\\
							 0.1 &  \textbf{65.50}& \textbf{88.44}\\
							0.2 & 64.89&88.03\\
                                0.5 & 64.08&87.34\\
                                
							\bottomrule
						\end{tabular}}
      \hspace{0.017\linewidth}
\subfigure[{Hyper-parameter $N$}
                        \label{tab:ablation_softlabel}]{
						  \tablestyle{4pt}{1.05}

    						\begin{tabular}{L{2cm}C{1.3cm}C{1.3cm}}
							\toprule
                                $N$&\makecell[c]{Top-1\\Accuracy}&\makecell[c]{Top-5\\Accuracy} \\
							\midrule
							5 & 63.75 & 86.93\\
							 10 & \textbf{63.99} & \textbf{87.61} \\
							20 & 63.74 & 87.45\\
							50 & 63.79&87.20\\
							100& 63.88&87.52\\
							\bottomrule
						\end{tabular}}
      \hspace{0.017\linewidth}
                \subfigure[{Combination of Perturbations}
                        \label{tab:ablation_pipeline}]{
							\tablestyle{4pt}{1.05}
    						\begin{tabular}{L{2cm}C{1.3cm}C{1.3cm}}
							\toprule
                                \makecell[l]{Strategy} &\makecell[c]{Top-1\\Accuracy}&\makecell[c]{Top-5\\Accuracy}\\
							\midrule
                                Baseline & 63.39&87.59\\
							Serial & 64.54 &86.92\\
							Parallel & 65.16 & 88.35\\
							 Random Pick & \textbf{65.50}& \textbf{88.44}  \\
                                && \\
							\bottomrule
						\end{tabular}}

	\end{center}
         \caption{RIS ablation experiments on CIFAR-100 with ResNet20. Our default settings are marked in gray.}
         \label{tab:weight_components}
\end{table*}

\subsubsection{Study on Different Input/Weight Perturbation Strategies} \label{compare_VDA}
Recall in RIS, We introduce perturbations on model input and weight for robustness modeling. Now, we investigate the impact of different perturbation strategies on performance. The results are summarized in Tab. \ref{tab:ablation_input} and \ref{tab:ablation_weight}. For input perturbations, we perform various data augmentation approaches on synthetic images, it's observed in Tab. \ref{tab:ablation_input} that all strategies outperform the baseline, which implies the versatility of RIS. In practice, we randomly choose input perturbation for each training batch to prevent overfitting.
For weight perturbations, we evaluate the effectiveness of adding Gaussian noise, adversarial disturb, and Dropout in Tab. \ref{tab:ablation_weight}, which indicates that different methods can bring noticeable performance gains. In subsequent experiments, we add Gaussian noise to the model parameters as the weight perturbation because it has low computational overhead and the best performance.

\subsubsection{A Closer Look at Diversity-aware Synthesis} 
\label{hyper-parameter of softlabel}
\begin{figure}[h]
	\centering
        \subfigure[GDFQ (baseline)]{
        \label{original_image}
		\includegraphics[width=0.48\linewidth]{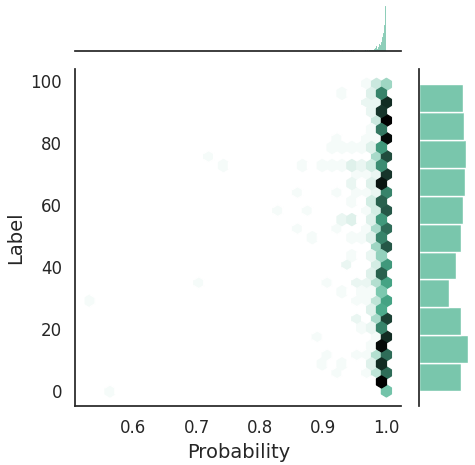}}
	\subfigure[RIS (ours)]{
		\label{softlabel_image}
		\includegraphics[width=0.48\linewidth]{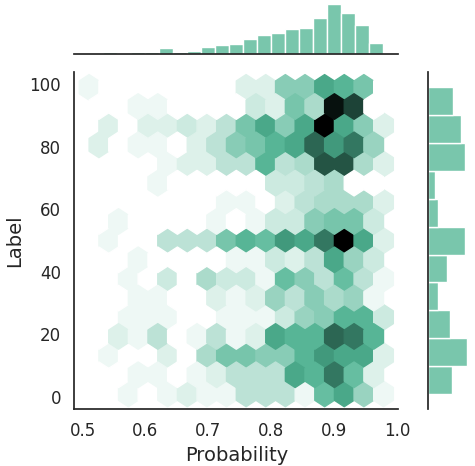}}
	\label{fig:cosine}
	\caption{Visualization of teacher predictions on synthetic images.} 
	\label{fig:softlabel}
\end{figure}

To verify that the proposed RIS can generate more diverse images, we first generate 1000 images using the original generator and the generator driven by soft labels, respectively. Then, we put these two sets of images into the teacher model and document their predicted labels and max probabilities, as visualized in Fig.\ref{fig:softlabel}. It is evident that the generator using labels with minimal correlations can synthesize a wider variety of images with more distinct differences between them, while the baseline method is prone to produce similar images. 

\subsubsection{Effectiveness of Each Component}
In RIS, we design the robustness objective for the generator via the inconsistency before and after input/weight perturbations and propose the diversity-aware scheme which involved soft labels with minimal correlations to avoid homogenization.
Table.~\ref{tab:ablation_component} documents the top-1 accuracy when adding different parts of the RIS method, from which it can be observed that every single strategy is able to boost the model precision individually. By applying input perturbation to the images generated by the generator and weight perturbation to the full-precision model, both types of disturbance can improve the accuracy of the quantized model.
And applying the pre-computed soft labels also contributes to improving the performance of the quantized model by instructing the generator to synthesize images with small correlations. When combining all components together, RIS achieves the largest performance gain.

\subsubsection{Changing of Hyperparameters}
\begin{figure}[h]
	\centering
        \subfigure[Study on Perturbations]{
        \label{perturbations correlation}
		\includegraphics[width=0.48\linewidth,height=0.4\linewidth]{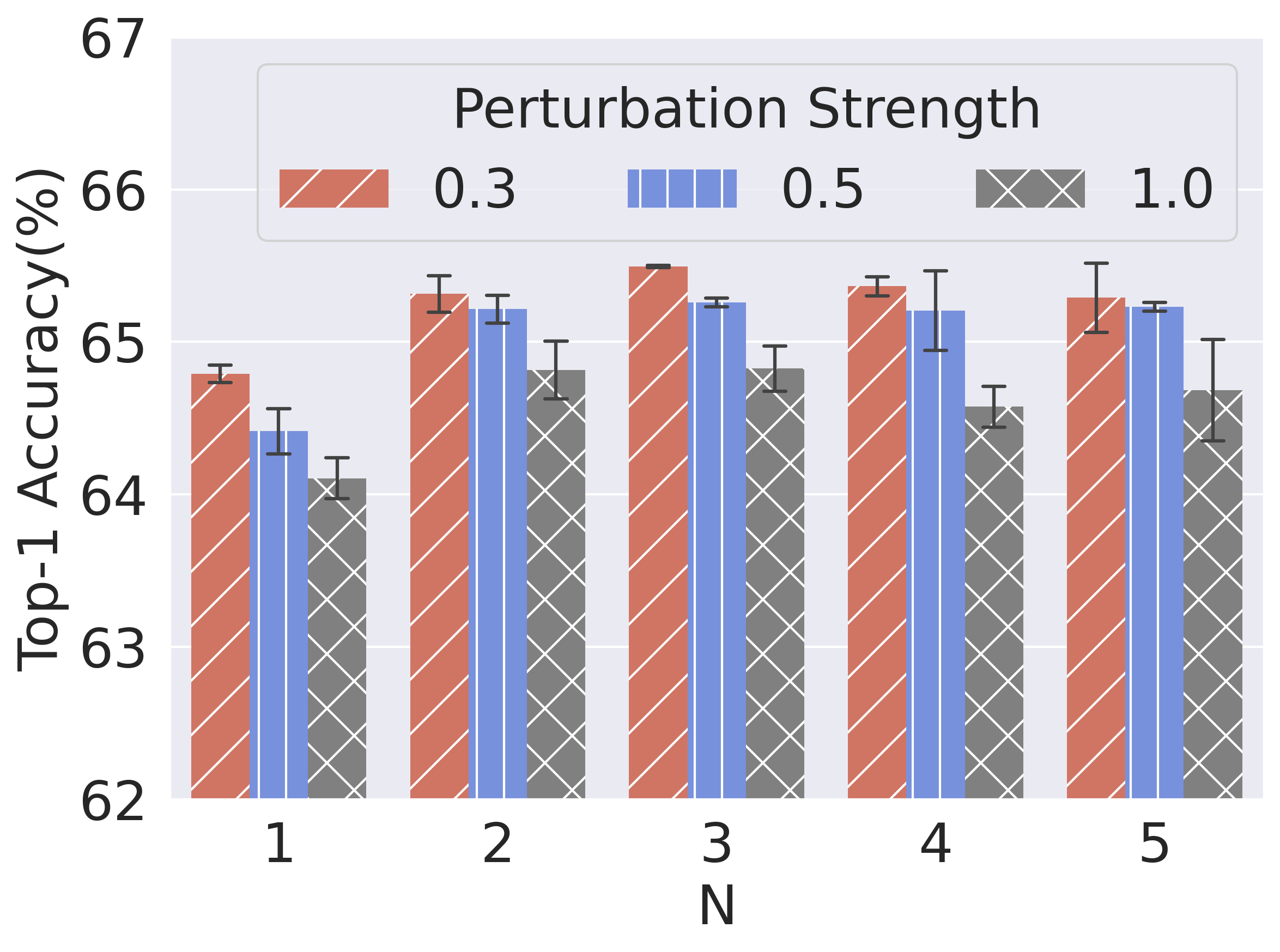}}
	\subfigure[Training Strategies]{
		\label{strategies correlation}
		\includegraphics[width=0.48\linewidth,height=0.4\linewidth]{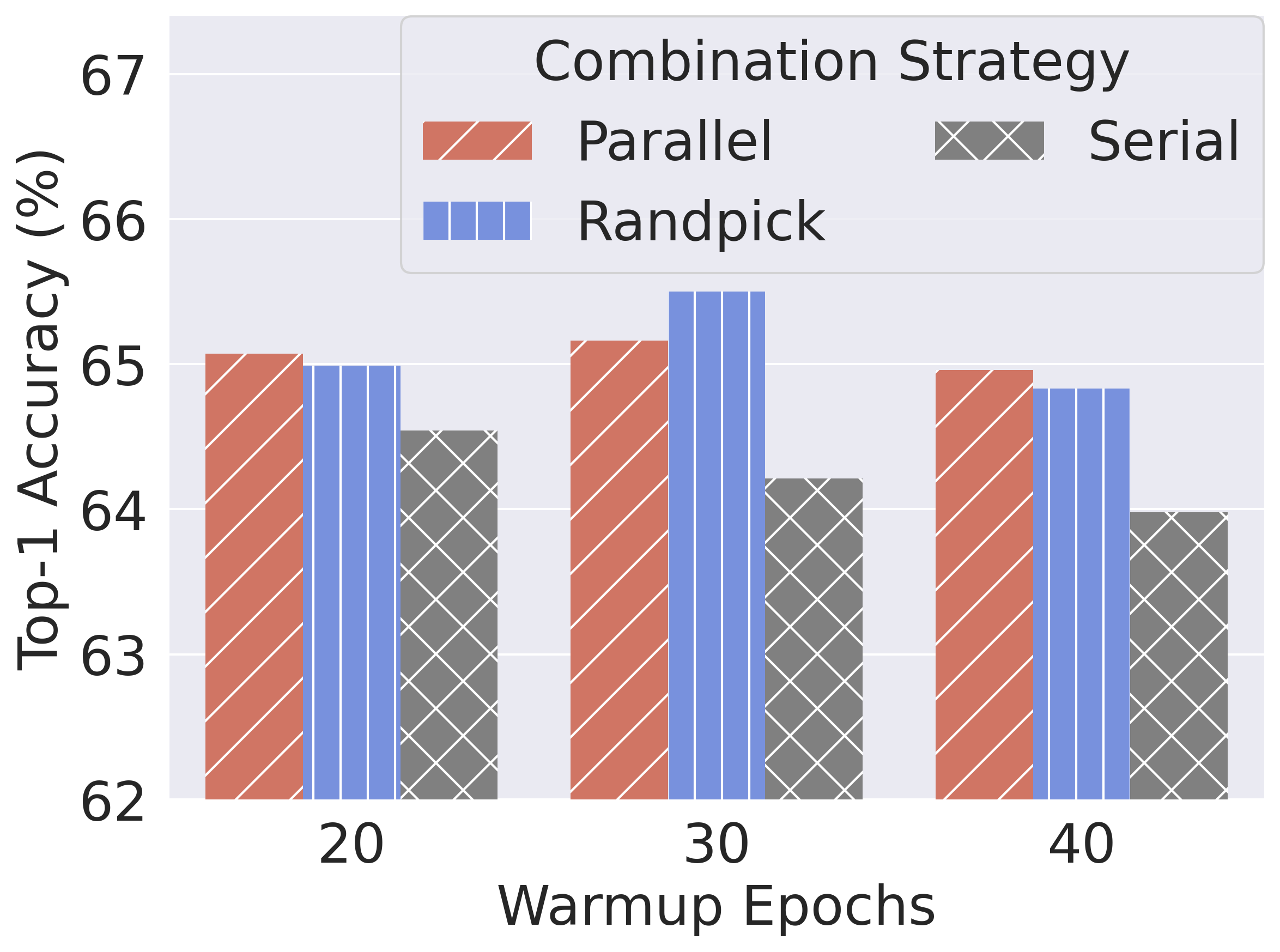}}
	\label{fig:cosine}
	\caption{Comparison of different hyper-parameters.}
	\label{fig:hyperparameters}
\end{figure}

We also show the effect of hyper-parameters involved in RIS. Tab. \ref{tab:ablation_softlabel} shows the empirical results of changing soft label number $N$. We observe similar accuracy gains when scaling $N$. Tab.~\ref{tab:ablation_epsilon} shows the classification accuracy when changing the threshold percentile $\epsilon$ in Eq. \ref{equ:theta}. The limited fluctuations in performance prove that RIS is robust to hyper-parameters. 
Fig. \ref{fig:hyperparameters} further presents the correlation between various hyperparameters. From Fig. \ref{perturbations correlation} it can be observed that the performance improves with the increase of N and remains stable after $N\geq3$ meanwhile small weight perturbation strength is most beneficial to the model. Fig. \ref{strategies correlation} shows the performance of warmup epochs and combination strategy demonstrating that suitable warmup epochs and combination strategy can effectively improve model performance.


\subsubsection{Study on Combination Strategy of Perturbations}
Since there are two types of different perturbations in our method, they can be calculated in three combinations which are presented in Tab. \ref{tab:ablation_pipeline}. Serial means putting the perturbed images to the perturbed model for computing $\mathcal{L}_{\text{robust}}$, which increases the accuracy by 0.37\%. While parallel refers to feeding the augmented images to the original model meanwhile giving the original image to the perturbed model, and then the losses of the two parts are added as $\mathcal{L}_{\text{robust}}$. This strategy can boost the performance by 1.57\% but less than random pick which increases the accuracy significantly by 2.11\% that selects a perturbation method between input perturbation and weight perturbation with equal probability.
	
\subsection{Quality Analysis on Synthetic Images}
We use FID \cite{FID} and IS \cite{IS} to evaluate the quality of the synthesized images at the statistical level. Both are well-known criteria in the field of GAN. As shown in Table~\ref{tab:fid}, our method outperforms GDFQ by 54.94 in FID and 9.86 in IS on ResNet-20, CIFAR-100. 
Our method results in synthesizing images with higher visual fidelity and more distinctive category-related features.

\begin{table}[!h]
	\begin{center}
        
		\label{tab:fid}

		\begin{tabular}{L{1.5cm}C{1.2cm}C{1.4cm}C{1.2cm}C{1.4cm}}
			\toprule
			\multirow{2}{*}{Method} & \multicolumn{2}{c}{CIFAR-10} & \multicolumn{2}{c}{CIFAR-100} \\ \cmidrule(r){2-3} \cmidrule(r){4-5}
			& \multicolumn{1}{c}{IS} & FID & \multicolumn{1}{c}{IS} & FID\\ \midrule
			GDFQ   &\multicolumn{1}{c}{3.96} & 120.91                & \multicolumn{1}{c}{1.95}            & 142.42\\ 
			DSG    &\multicolumn{1}{c}{3.96}  &324.84                  & \multicolumn{1}{c}{2.48}            & 184.27\\ 
			Qimera   &\multicolumn{1}{c}{2.02}  & 145.89               & \multicolumn{1}{c}{2.19}            & 130.87\\ 
			RIS    &\multicolumn{1}{c}{\textbf{10.4}}  & \textbf{97.45}                & \multicolumn{1}{c}{\textbf{11.81}}            & \textbf{87.48}\\ \bottomrule
		\end{tabular}
	\end{center}
 		\caption{The FID \cite{FID} and IS \cite{IS} of the synthetic images.}
        \label{tab:fid}
\end{table}

\section{Conclusions and Limitations}

In this paper, we propose Robustness-Guided Image Synthesis (RIS) to improve the quality of synthetic images in data-free scenarios. Specifically, we propose to explicitly model image robustness on the basis of inconsistency at feature and prediction levels, and design a robustness-guided scheme that enables the generator to synthesize images with both rich semantics and diversity-aware. We conduct an extensive set of experiments, showing that RIS outperforms various existing data-free quantization methods, and can be further extended to data-free knowledge distillation. Nevertheless, there are some limitations. Firstly, how to generalize our approach to methods without a generator is worth exploring. Secondly, how to explicit numerous OOD data in the wild is needed to investigate in future work.
\clearpage

\section{Acknowledgements}
We thank the anonymous AAAl reviewers for providing uswith valuable feedback that greatly improved the qualityof this paper. Jianhong Bai would also like to thank Huan Wang from Northeastern University (Boston, USA) for his selfless guidance and help.

This work is supported by the Zhejiang Provincial Key RD Program of China (Grant No. 2021C01119) and the National Natural Science Foundation of China (Grant No. U21B2004, 62106222), the Natural Science Foundation of Zhejiang Province, China (Grant No. LZ23F020008) and the Zhejiang University-Angelalign Inc. R$\&$D Center for Intelligent Healthcare.
\bibliography{aaai24}

\clearpage

\end{document}